\documentclass[runningheads]{llncs}
\usepackage{algorithm}
\usepackage{algorithmic}
\usepackage{mathrsfs}
\def\eg{\emph{e.g.}}

\def\etal{\emph{et al.}}
\def\ie{\emph{i.e.}}

\usepackage{booktabs}
\usepackage{multirow}

\usepackage{graphicx}
\usepackage{amsmath,amssymb} % define this before the line numbering.
\usepackage{color}
\usepackage[width=122mm,left=12mm,paperwidth=146mm,height=193mm,top=12mm,paperheight=217mm]{geometry}
\begin{document}
% \linenumbers
\pagestyle{headings}
\mainmatter

\title{Searching Action Proposals via Spatial Actionness Estimation and Temporal Path Inference and Tracking} % Replace with your title

\titlerunning{Searching Action Proposals}

\authorrunning{$Nannan \hspace{2pt}Li, Dan \hspace{2pt}Xu, Zhenqiang \hspace{2pt}Ying, Zhihao \hspace{2pt}Li, Ge \hspace{2pt}Li$}

\author{$Nannan \hspace{2pt}Li^{\sharp}, Dan \hspace{2pt}Xu^{\S}, Zhenqiang \hspace{2pt}Ying^{\sharp}, Zhihao \hspace{2pt}Li^{\sharp}, Ge \hspace{2pt}Li^{\sharp}$}

%Please write out author names in full in the paper, i.e. full given and family names.
%If any authors have names that can be parsed into FirstName LastName in multiple ways, please include the correct parsing, in a comment to the volume editors:
%\index{Lastnames, Firstnames}
%(Do not uncomment it, because you may introduce extra index items if you do that...)

\institute{$^{\sharp}$Peking University Shenzhen Graduate School, Shenzhen, P.R.China \\ $^{\S}$DISI, University of Trento, Trento, Italy \\	
	%\email{ \{linn\}@pkusz.edu.cn}
}

\maketitle

\begin{abstract}
In this paper, we address the problem of searching action proposals in unconstrained video clips. Our approach starts from actionness estimation on frame-level bounding boxes, and then aggregates the bounding boxes belonging to the same actor across frames via linking, associating, tracking to generate spatial-temporal continuous action paths. To achieve the target, a novel actionness estimation method is firstly proposed by utilizing both human appearance and motion cues. Then,
%Following~\cite{yu2015fast},
the association of the action paths is formulated as a maximum set coverage problem with the results of actionness estimation as a priori. To further promote the performance, we design an improved optimization objective for the problem and provide a greedy search algorithm to solve it. Finally, a tracking-by-detection scheme is designed to further refine the searched action paths.
Extensive experiments on two challenging datasets, UCF-Sports and UCF-101, show that the proposed approach advances state-of-the-art proposal generation performance in terms of both accuracy and proposal quantity.
\end{abstract}

\section{Introduction}
Video action analysis is an important research topic for human activity understanding, and has gained a wide attention in recent years. A common task of video action analysis is action recognition, which aims to identify which type of action is occurring in a video volume~\cite{wang2011action,wang2013action,simonyan2014two}. Compared to action recognition, action detection is a more difficult task, as it requires not only determining the action class, but also localizing the action in the video. Similar to the object detection task, in which reliable object proposals play a crucial role in the detection performance~\cite{uijlings2013selective}, action proposal is also a fundamental problem in action detection.

This paper focuses on generating high-quality action proposals in both spatial compactness and temporal continuity. Existing works in the literature have made different efforts to address the problem, including segmentation-and-merging strategy~\cite{ma2013action,oneata2014spatio,bergh2013online,jain2014action}, dense motion features~\cite{weinzaepfel2015learning,wang2013dense,van2015apt}, human-centric models~\cite{prest2013explicit,klaser2010human}, and object proposals based approaches~\cite{yu2015fast,gkioxari2015finding}. Despite promising results achieved in these works, video action proposal generation is still a challenging problem due to the complex spatio-temporal relationship modeling involved in the task. The problem can be considered as a task with two essential steps, namely spatial (\ie frame-level) actionness estimation and temporal (\ie video-level) action path generation. For one aspect, because of the large diversity and variation of human actions, it is difficult to generate robust frame-level actionness proposals which contain meaningful action motion patterns and are clearly discriminative from the background in unconstrained videos. For the other aspect, as a fact that the whole number of potential actionness regions on each frame usually has an exponential growth of the video duration~\cite{tran2014video}, it is extremely impractical to calculate on all possible connections of the regions for generating the action paths and guaranteeing each of them associated with the same actor(s).

To tackle the above mentioned issues in the generation of action proposals, we propose a novel framework based on spatial actionness estimation from multiple cues and temporal action path extraction from a fast inference and tracking. Firstly, unlike previous works using selective search~\cite{uijlings2013selective} or edge boxes~\cite{zitnick2014edge} for generating object proposals for actionness estimation, we employ more action related cues including both human and motion. Secondly, a deep Faster-RCNN~\cite{ren2015faster} network is trained and fine-tuned on augmented action detection datasets for obtaining accurate human proposals. Then action motion patterns with Gaussian Mixture Models are modeled for motion estimation of each human proposal. Both human and motion estimations are feed into a proposed forward and backward search algorithm for video-level action path generation. Finally, we use a tracking-by-detection approach to refine the action path by supplement actionness proposals missing in frames.

The key contributions of this paper include three folds: i) we construct an action detector at frame-level by taking both appearance and motion clues into account, which can handle the problem of detecting human with uncommon poses and discriminate actionness proposals containing meaningful motion patterns from the backgrounds; ii) We formulate the action path generation as a maximum convergence problem \cite{nemhauser1978}. We propose an improved optimization objective for the problem and provide a greedy search algorithm to solve it. iii) Extensive experiments on UCF-Sports, UCF-101 datasets show that the proposed method achieves the state-of-the-art performance compared with other existing approaches.

\begin{figure}[!t]
\includegraphics[width=1.00\textwidth,height=0.55\textwidth]{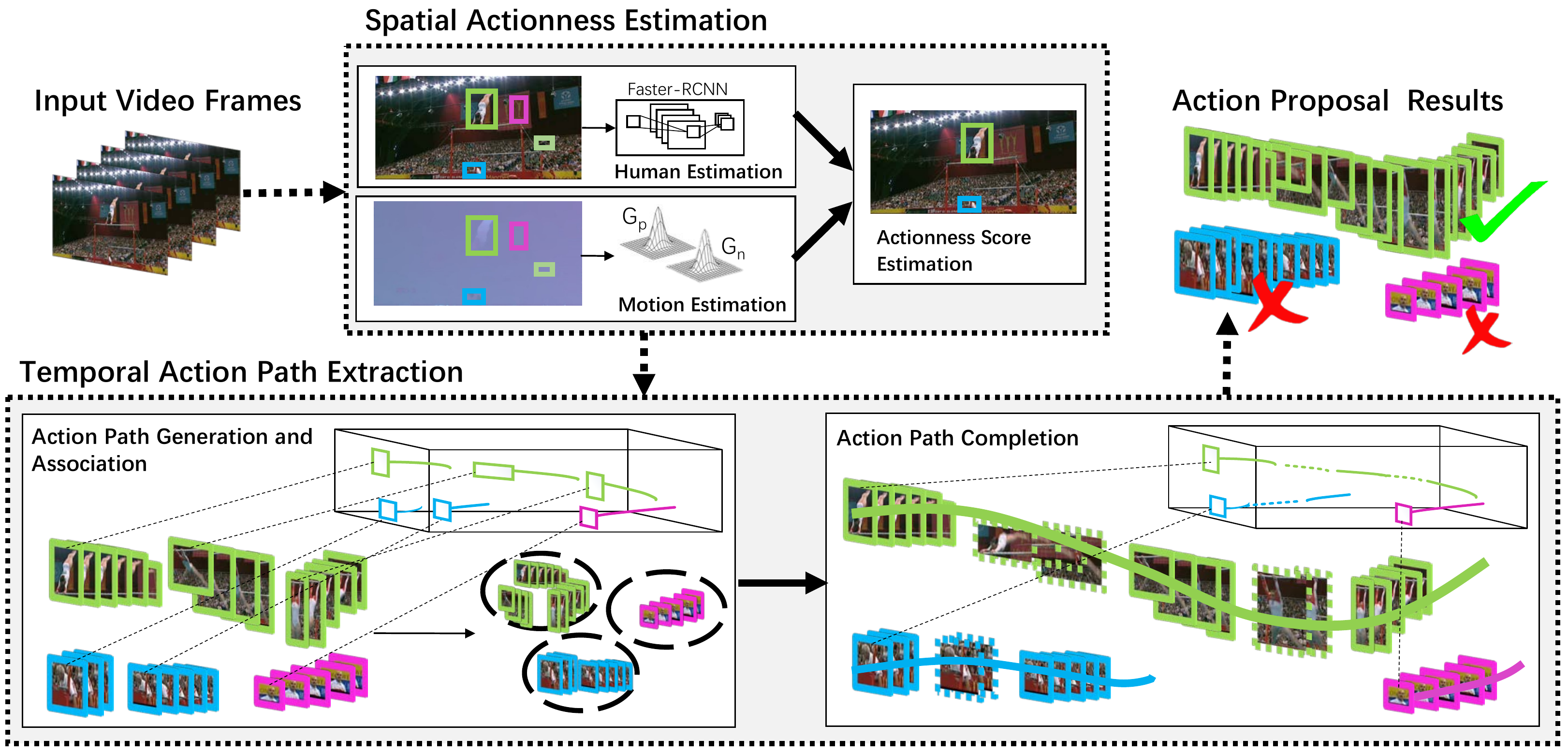}
\caption{The framework of the proposed action proposal generation approach.}
\end{figure}

\section{Related Work}
Traditionally, action localization or detection is performed by sliding window based approaches \cite{siva2011weakly,laptev2007retrieving,gaidon2013temporal,wang2015robust}. For instance, Siva \etal \cite{siva2011weakly} proposed a supervised model based on multiple-instance-learning to slide over subvolumes both spatially and temporally for action detection.
Instead of performing an exhaustive search through sliding over the whole video volumes, Oneata \etal \cite{oneata2014spatio} put forward a branch-and-bound search approach to achieve the time-efficiency.
The main limitation of these sliding-window based approaches is that the detection results are confined by a video subvolume, and thus can not accurately capture the varying shape of the motion.

Some research works address the problem by employing segmentation-and-merging strategy. Generally,
these methods include three steps: i) segment the video; ii) merge the segments to generate tube proposals; iii) represent tubes with dense motion features and construct action classifier for recognition. For instance, in \cite{jain2014action} action tubes are generated by hierarchically merging super-voxels. However, accurate video segmentation is a difficult problem
% and it is difficult to efficiently segment the human action from videos with
especially under unconstrained environments. To alleviate the difficulty encountered with segmentation, some other methods use a figure-centric based model. In \cite{prest2013explicit} the human and objects are detected first and then their interactions are described.
Kl{\"a}ser \etal  \cite{klaser2010human} detect human on each frame and track the detection results across frames using optical flow. Our approach also utilizes tracking, via a more robust tracking-by-detection approach \cite{hare2011struck,kalal2012tracking} based on a combined feature representation of color and shape.

Recently, some methods built upon generation of action proposals are presented. Gkioxari \etal \cite{gkioxari2015finding} proposed to utilize Selective Search method for proposing actions on each frame, then scored those proposals by using features extracted by a two-streams Convolutional Neural Networks (CNN), and finally, linked them to form ation tubes.
Philippe \etal \cite{weinzaepfel2015learning} adopted the same feature extraction procedure, then utilized a tracking-by-detection approach to link frame-level detections, in combination with a class-specific detector. Our method replaces object proposal method and two-stream CNN with the Faster R-CNN model for calculation efficiency.
%employs Faster R-CNN model for generating action proposals, replacing object proposal method and two-streams CNN for calculation efficiency.
The most related work to ours is that presented in \cite{yu2015fast}, in which actionness score is calculated for each action path and then a greedy search method is used to generate proposals.
% The major differences between our work and theirs exist in three aspects:
Our work differentiates from theirs in the following three aspects:
i) we train a Faster R-CNN model for human estimation, which has a stronger ability to differentiate human from backgrounds; ii) compared with the optimization objective they proposed, our improved optimization objective simultaneously maximizing actionness score and member similarity in a path set, thus can effectively cluster the paths from the same actor into a group; iii) we utilize a tracking-by-detection approach to supplement the missing detections.

\section{The Proposed Approach}
The proposed approach takes video clip as input and generates action proposal results. The framework of our approach is illustrated in Fig. 1. The main procedure consists of two stages: spatial actionness estimation and temporal action path extraction. Firstly, bounding boxes at frame-level that may contain meaningful motion are extracted by simultaneously considering appearance and motion cues; then action paths corresponding to the same actor at video-level are generated and linked to obtain action proposals. The details of our method will be elaborated in the following sections.

\subsection{Spatial Actionness Estimation} \label{sec:spatial_action}
\subsubsection{Human Estimation}\label{humanesti}
\begin{figure}[!t]
\centering
\includegraphics[width=1.0\textwidth,height=0.17\textwidth]{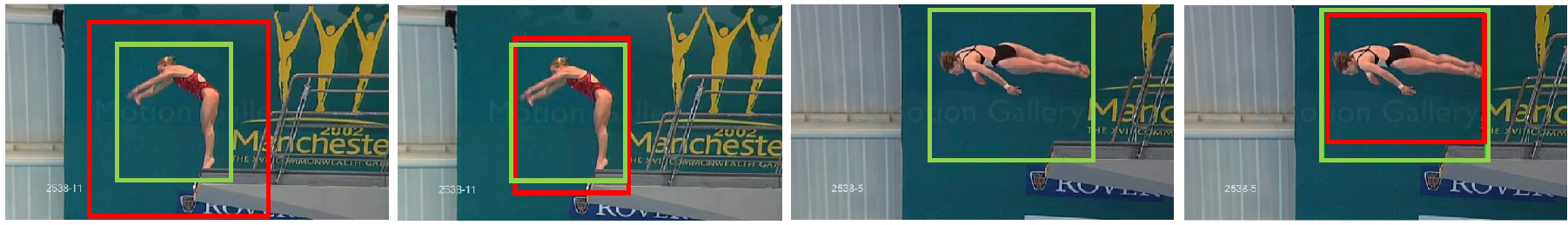}
\caption{ Comparison of human detection results. The bounding boxes with red and green color are the ground truth and the detection results respectively. The 1-st and 3-nd columns are from the initial Faster R-CNN~\cite{ren2015faster} (There is a missing detection in the 3-nd column); while the 2-nd and 4-th columns are from our fine-tuned model.}
%Notice that the detection misses hitting in the third column.}%\tode{add a second line of figures corresponding to the first line with motion proposal bounding boxes and scores.}}
\label{vsfaster-r-cnn}
\end{figure}

Detection of Human proposal is an important and heuristic step for action localization. We implement the human proposal detection employing the Faster R-CNN~\cite{ren2015faster} pipeline with a VGG-16 model~\cite{simonyan2014very} pre-trained on ILSVRC dataset~\cite{russakovsky2015imagenet}. Faster R-CNN introduces a Region Proposal Network (RPN) that simultaneously predicts object bounding boxes and their corresponding objectness scores in near real-time speed.
As the human detection task is a binary-classification problem, the output of the classification layer of Faster-RCNN network is revised to a two-way softmax classifier: one for the `human' class and the other for the `background' class.
For action classes such as diving and swing, the appearance (especially for the shape and the pose) of the human changes significantly among the whole action duration.
Therefore, the detection network fine-tuned on the standard PASCAL VOC 2007 dataset is unable to effectively detect the human under those circumstance.
To handle the problem, we perform a data augmentation by merging the training data of the human class of PASCAL VOC 2007 and 2012, and rotating each training sample with seven different angles from $\frac{\pi}{4}$ to $\frac{7\pi}{4}$ with an interval of $\frac{\pi}{4}$.
Let $b^i_t$ denotes the bounding box for the $i$-th human proposal at $t$-th frame.
The bounding box is represented as $[x,y,w,h]$, where $w$ and $h$ stand for width and height respectively, and $(x,y)$ is the center.
After training, for each bounding box $b_*^*$ in the test video,
a probability $S_h(b_*^*)$ can be estimated by the CNN network.
By setting a probability threshold, human proposals with higher probability are kept for follow-up processing. A comparison of human detection results between original Faster R-CNN model and our refined one is showed in Fig. \ref{vsfaster-r-cnn}, from which it can be clearly observed that detection results from refined model are more precise and compact.

\subsubsection{Motion Estimation}
Human cue provides important prior information for generating frame-level action proposals, however it is not sufficient to determine whether an action occurs, \eg, human standing still. Thus we propose to further utilize motion cue for discarding false positive action proposals.
%For each training video frame, we obtain candidate human proposals from RPN of Faster-RCNN. The proposals with an intersection-over-union (IoU) overlapping with the groundtruth bounding box more than 0.7 are selected as positive (\ie action) proposals, and those with the IoU overlapping less than 0.1 as negative (\ie background) ones.
The histograms of optical flow (HOF) \cite{wang2013dense} descriptor is used to describe the motion pattern of each human proposal. We construct two Gaussian Mixture Models (GMMs) $G_p(.)$ and $G_n(.)$ upon the HOFs, which represent the positive and negative proposal class respectively, to predict the probability of a motion pattern belonging to the actions or the background. HOFs calculated within bounding boxes of an Intersection-over-Union (IoU) overlapping with  ground truth more than $0.5$ are used as positive samples, while those with IoU overlapping less than $0.1$ as negative samples.
Given a test proposal $b^i_t$ and its HOF $h_i$, we define
the likelihood of $b^i_t$ being a motion score using the predictions from two mixture of Gaussian models as:
\begin{equation}\label{motionscore}
   S_m(b^i_t) = \sigma(G_p(h_i)/G_n(h_i)),
\end{equation}
where $\sigma = 1/(1+e^{-x})$ maps likelihood into the range [0, 1]. To reduce the influence induced by camera movement to optical flow calculation, we adopt the approach presented by \cite{wang2013action} to estimate camera motion and subtract it to obtain robust optical flow.

\subsubsection{Actionness Score Calculation}

The actionness score of a bounding box $b^i_t$ consists of two parts: human detection score and motion score, and is defined as follows:
\begin{equation}\label{action-score}
    S(b^i_t) = S_h(b^i_t) + \lambda_p *S_m(b^i_t),
\end{equation}
where $\lambda_p$ is the parameter that balances the human estimation and motion estimation score.

\subsection{Temporal Action Path Extraction}

\subsubsection{Problem Formulation}

Given action proposals on each frame, our goal is to find a set of action paths $\textbf{P} =\{p_1, p_2, \ldots, p_i\}$, where $p_i = \{b^i_s, b^i_{s+1},\ldots, b^i_e\}$ corresponds to a path that starts from $s$-th frame and ends at $e$-th frame.
 Yu and Yuan \cite{yu2015fast} formulate finding action path set $\textbf{P}$ as a maximum set coverage problem (MSCP) and propose an optimization objective maximizing actionness score. Inspired by their work, we formulate it as a MSCP with an improved optimization objective, which simultaneously maximizes actionness score and similarity among members within the path set $\textbf{P}$.
Formally, our optimization objective can be presented as follows:

\begin{equation}\label{MHSCP}
\begin{split}
& \max\limits_{\textbf{P}\in\Phi} \sum\limits_{b_t\in \bigcup p_i} S(b_t) + \sum\limits_{i,j}W(p_i, p_j) \\
& s.t. \quad \space\space  |\textbf{P}| \leq N \\
&   \quad \quad {\textbf{O}(p_i, p_j) } \leq \eta_p, \forall p_i, p_j \in \textbf{P}, i\neq j  ,
\end{split}
\end{equation}
where $W(p_i,p_j)$ represents the similarity between action path $p_i$ and $p_j$, and its definition will be explained in subsection action-path-association; $S(b_t)$ is the actionness score of bounding box $b_t$ (cf. Eq. \ref{action-score});
$\Phi$ is action-path-candidate set; $\eta_p$ is a threshold. The first constraint in Eq. \ref{MHSCP} sets the maximum number of paths contained in \textbf{P}; while the second constraint facilitates \textbf{P} to avoid generating redundant action paths that are overlapped. The overlapping of two paths is evaluated by $\textbf{O}(p_i,p_j)$, which is defined as follows:

\begin{equation}\label{tubeoverlap}
    \textbf{O}(p_i, p_j) = \frac{1}{\max(t^i_e,t^j_e)-\min(t^i_s,t^j_s)} \bullet \sum\limits_{\max(t^i_s,t^j_s)\leq t \leq \min(t^i_e,t^j_e)}o(b^i_t,b^j_t)  .
\end{equation}
In Eq. \ref{tubeoverlap}, $o(b^i_t,b^j_t)$ is defined as $\frac{\cap(b^i_t,b^j_t)}{\cup(b^i_t,b^j_t)}$, representing for IoU of two bounding boxs $b^i_t$ and $b^j_t$.

\subsubsection{Action Path Generation}
To solve the MSCP in Eq. \ref{MHSCP}, the action-path-candidate set $\Phi$ needs to be obtained first. We wish that $\Phi$ consists of spatio-temporal smooth path $p_i$ whose consecutive elements $b^i_t, b^i_{t+1}$ should satisfy the following two requirements:

\begin{equation}\label{link_box}
  \begin{split}
& \hphantom{PPPPPPPPPPPPPPPPPP} o(b^i_t,b^i_{t+1}) \geq \eta_o \\
& \|C(b^i_t)-C(b^i_{t+1})\| + \lambda_a \|H(b^i_t)-H(b^i_{t+1})\| \leq \eta_f  ,
  \end{split}
\end{equation}
where $o(b^i_t,b^i_{t+1})$ represents IoU, as defined in Eq. \ref{tubeoverlap}; $C(b^i_t)$ and $H(b^i_t)$ stand for histograms of color (HOC) and histograms of gradient (HOG) of $b^i_t$, and $\lambda_a$ is a trade-off balancing the weight of the two terms; $\eta_o$ and $\eta_f$ are thresholds. The first requirement in Eq. \ref{link_box} ensures that consecutive bounding box $b^i_t$ and $b^i_{t+1}$ are spatially continuous; the second requirement ensures that $b^i_t$ and $b^i_{t+1}$ have similar appearance, thus the path $p_i$ may follow the same actor.

To obtain $\Phi$, we adopt the method proposed by \cite{yu2015fast} with minor modification to avoid generating much highly-overlap paths. The algorithm includes two stages: forward search and backward track. The aim of the former is to locate the end of a path; while that of the latter seeks to recover the whole path.
 The central idea is to maintain an updating pool of best Top-$N$ path candidates, which is represented as $\Phi = {(\tau_k, b^k ), k=1,2,\ldots,N}$, where $\tau_k$ is the score of path $k$ and obtained by
accumulating $S(b^k_t)$ of $b^k_t$ it passes by; $b^k$ is the bounding box of the end of $k$-th path. In the forward search, it also records an accumulated actionness score of each $b^i_t: \tau(b^i_t) = \max\limits_{b^j_{t-1}}{\tau(b^j_{t-1})+S(b^i_t)}$,where $b^j_{t-1}$ and $b^i_t$ satisfy the two requirements in Eq. \ref{link_box}. Given $b^i_t$ at frame $t$, we update path candidate pool according to the following two steps: first, for each candidate $(\tau_k, b^k ), k=1,2,\ldots,N$, if there exists any $b^i_t$ connecting to $b^k$, then $b^k$ will be replaced by $b^i_t$ that has the largest $\tau(b^i_t)$; second, if the accumulated score of $b^i_t$ is larger than the score of $N$-th proposal, i.e. $\tau(b^i_t)>\tau_N$, then $(\tau_N, b^N )$ is updated as $(\tau(b^i_t), b^i_t)$. After the forward search, a backward trace is performed to recover each $b^k_t$ on the candidate path $(\tau_k, b^k )$. More specifically, for path $p_k$, we obtain $\{b^k_t: t=t_s, t_{s+1}, \ldots,t_e\}$ by solving the equation: $\tau_k = \sum\limits_{t_s\leq t \leq t_e}S(b^k_t)$.

The pseudo-code of forward-backward search is illustrated in Algorithm \ref{forward-backward}. It takes bounding box score $S(b^i_t)$ as input data and outputs action paths $p_k, k=1,2,\ldots,N$. The lines 1 to 8 describe forward search and line 9 corresponds to backward trace. In line 3, $N^b_t$ denotes the number of bounding box on frame $t$.

    \renewcommand{\algorithmicrequire}{\textbf{Input:}}
    \renewcommand{\algorithmicensure}{\textbf{Output:}}
      \begin{algorithm}[t]
        \caption{\quad \quad \quad Forward Search and Backward Track}
        \begin{algorithmic}[1]
          \REQUIRE  bounding box score $S(b^i_t)$
          \ENSURE  action path $p_k,k=1,2,\ldots,N$
          \STATE $ \tau_k = 0,b^k = \emptyset, k = 1,2,\ldots,N $
          \FOR{$t=1 \to T$}
          \FOR{$i=1 \to N^b_t$}
          \STATE $\tau(b^i_t) = \max_{b^j_{t-1}}\tau(b^j_{t-1})+S(b^i_t)$
          \ENDFOR
          \STATE step1: update each candidate $(\tau_k, b^k)$ with $b^i_t$ that connects with $b^k$ and has the largest score $\tau(b^i_t)$
          \STATE step2: update $(\tau_N, b^N)$ as $(\tau(b^i_t,),b^i_t)$, if $\tau(b^i_t)>\tau_N$
          \ENDFOR
          \STATE backward trace to locate $b^k_t, t=t_s, t_{s+1}, \ldots, t_e$ in $p_k$
        \end{algorithmic}
        \label{forward-backward}
      \end{algorithm}

\subsubsection{Action Path Association}
\label{sec-action_path_association}

Once obtaining $\Phi$, the MSCP in Eq. \ref{MHSCP} can be solved. According to \cite{nemhauser1978}, the maximum set coverage problem is NP-hard but a greedy-search algorithm can achieve an approximation ratio of $1-1/e$. Here, we present a greedy-search solution to address the problem. In the beginning, we search for the candidate $p_i$ with the largest action score $\tau_k$ in $\Phi$, then add it into path set \textbf{P}. Supposing that \textbf{P} has contained $k$ action paths, we enumerate the rest paths in $\Phi$ and find the one that maximizes the flowing equation as the $k+1$-path $p_i$:

\begin{equation}\label{Eq5}
    \arg\max\limits_i \sum\limits_{b\in p_i\cup p_1 \cup \ldots \cup p_{k}} S(b) + \sigma (1/k \cdot \sum\limits_{j=1,2,\ldots,k}W(p_i,p_j) )  .
\end{equation}

In Eq. \ref{Eq5}, $W(p_i,p_j)$ represents the similarity of action path $p_i$ and $p_j$, and is defined as:
$W(p_i,p_j) = 1/(\| C(p_i)-C(p_j) \| + \lambda_a \| H(p_i)-H(p_j) \|)$, where $C(p_*)$ and $H(p_*)$ represent the cluster centers of HOC and HOG of bounding boxes from path $p_*$ respectively. The larger value of $W(p_i,p_j)$, the more likely that the paths $p_i$ and $p_j$ follow the same actor. To reduce redundant paths in set \textbf{P}, the newly added path $p_i$ should satisfy the constraint in Eq. \ref{tubeoverlap}.

\subsubsection{Action Path Completion}

\begin{figure}[t]
\centering
\includegraphics[width=0.95\textwidth]{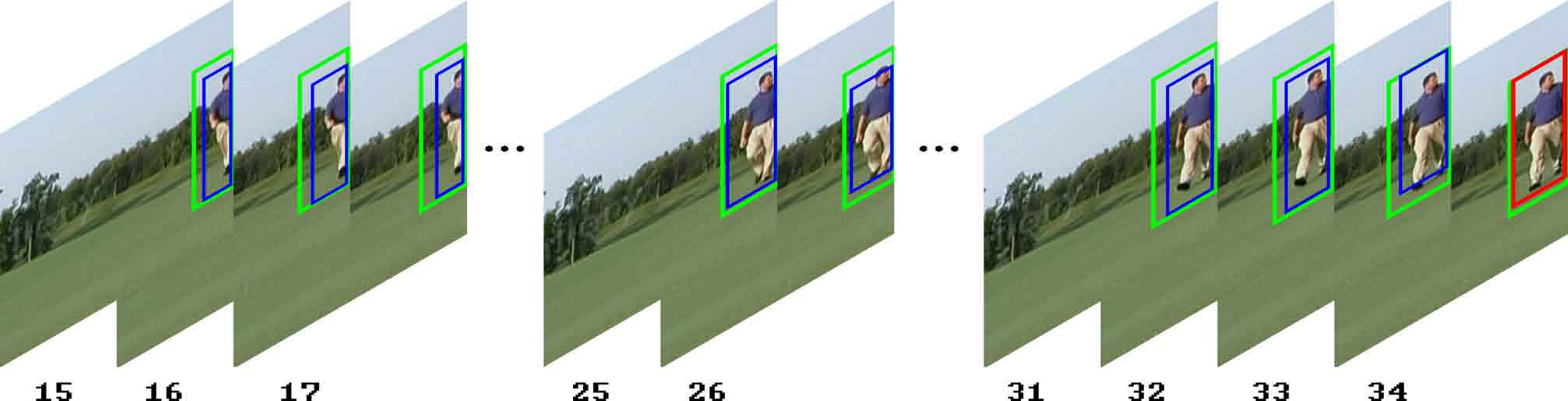}
\caption{Examples of tracking-by-detection results. The bounding boxes with green and red color are the groundtruth and the detected frame-level human bounding boxes respectively, and those with blue color are obtained by our tracking-by-detection strategy. All the missing human targets before frame $34$ are perfectly located by the tracking approach.}%\todo{Coud we put the figures flat?}}
\label{missing-detection}
\end{figure}

As human detection may miss hitting in some frames, the track obtained by connecting the paths in $\textbf{P}$ will have temporal gaps. To get a temporal-spatial continuous track of an actor, we fill the gaps by using tracking-by-detection approaches \cite{weinzaepfel2015learning}. We train a linear SVM as frame-level detector. The initial set of positives consist of bounding boxes in set $\textbf{P}$, while negatives compose of bounding boxes excluded from set $\textbf{P}$ and boxes that are randomly selected around positives with the IoU less than 0.3. Given the detection region $b_t$ on frame $t$, we intend to find the most likely location on frame $t+1$ where the human detection is missed. Firstly, we map $b_t$ to $b^{'}_{t+1}$ with the shift of the median of optical-flow inside region $b_t$; secondly, construct a search region $\overline{b^{'}_{t+1}}$ by extending the height and width of $b^{'}_{t+1}$ to half past one times of original length; thirdly,
scan $\overline{b^{'}_{t+1}}$ with a set of windows whose ratio between width and length varies in a range $[0.8,1.2]$ to adapt possible size change of an actor. The best region $b_{t+1}$ is selected as the one that maximizes the following equation: %$b_{t+1} = \arg\max\limits_{\gamma\in N(\overline{b^{'}_{t+1}})} S_f(\gamma)$,
\begin{equation}\label{tracking}
   b_{t+1} = \arg\max\limits_{\gamma \in N(\overline{b^{'}_{t+1}})} S_f(\gamma)  ,
\end{equation}
where $N(\overline{b^{'}_{t+1}})$ represents the window set produced by scanning $\overline{b^{'}_{t+1}}$ and $S_f(\cdot)$ is the SVM detector whose input feature is chosen as the combination of HOC and HOG. After obtaining $b_{t+1}$, we update the SVM detector by adding $b_{t+1}$ as a positive sample and boxes around $b_{t+1}$ with the IoU less than 0.3 as negatives. An example of how the tracking approach supplementing missing detections is illustrated in Fig. \ref{missing-detection}.

\subsection{Action Proposal Generation}
The spatio-temporal continuous track can be considered as an action tube that focuses on an actor from appearing until disappearing. For each action tube, if its duration is larger than a specified threshold ($e.g.$ 20), we regard it as an action proposal, denoted as $\mathscr{T}$.

\section{Experiment}
In this section, we describe the details of the experimental evaluation of the proposed approach, including datasets and evaluation metricts, implementation details, an analysis of the proposed approach and the overall performance comparison with state of the art methods.

\subsection{Datasets and Evaluation Metric}
We evaluate the performance of the proposed action proposal approach on two publicly available action-detection datasets: UCF-Sports \cite{ucfsports} and UCF-101 \cite{ucf101}.

\textbf{UCF-Sports} UCF-Sports dataset consists of 150 short videos of sports collected from 10 action classes. It has been widely used for action localization. The videos are truncated to contain a single action and bounding box annotation is provided for each frame.

\textbf{UCF-101} UCF-101 dataset has more than 13000 videos that belong to 101 classes.
In a subset of 24 categories, human actions are annotated both spatially and temporally. Compared with UCF-Sports, only a part of videos ($74.6\%$) are trimmed to fit the motion.

\textbf{Evaluation Metric} To evaluate the quality of the action proposal $\mathscr{T}$, we follow the metric proposed by \cite{van2015apt}.
More specifically, the estimation is based on the mean IoU value between action proposal $\mathscr{T}$ and ground truth $\textbf{G}$, which is defined as: $IoU(\textbf{G},\mathscr{T}) = \frac{1}{|\textbf{C}|}\sum\limits_{t\in\textbf{C}}o(\textbf{G}_t, \mathscr{T}_t)$, where $\textbf{G}_t$ and $\mathscr{T}_t$ are the detection bounding box and ground truth on t-$th$ frame respectively; $o(.,.)$ is the IoU value that is defined in Eq. \ref{tubeoverlap}; $|\textbf{C}|$ is the set of frames where either the detection result or the ground truth is not null.
An action proposal is considered as true-positive if $IoU(\textbf{G},\mathscr{T})\geq \eta$, where $\eta$ is a specified threshold. In the following passage, $\eta$ is set as 0.5 if not specified.

\subsection{Implementation Details}
The human estimation is implemented under the Caffe platform~\cite{jia2014caffe} and based on the Faster R-CNN pipeline with a VGG16 model for parameter initialization as described in Sec.~\ref{sec:spatial_action}.
We use a four-step alternating training strategy \cite{ren2015faster} to optimize two pipelines (\ie RPN and Fast RCNN) of the whole network. For training the RPN pipeline, the same settings of scales and aspect ratios are used as in~\cite{ren2015faster}. For training the Fast RCNN pipeline, the mini-batch size is set to 128, and the ratio of positive to negative samples is set to 1:4. The network is trained with Stochastic Gradient Descent (SGD) with an initial learning rate of 0.001 and drop by 10 times at every the 5-th epochs, and the momentum and weight decay are set as 0.9 and 0.0005 respectively.
\begin{figure}[t]
\centering
\includegraphics[width=0.62\textwidth, height=0.42\textwidth]{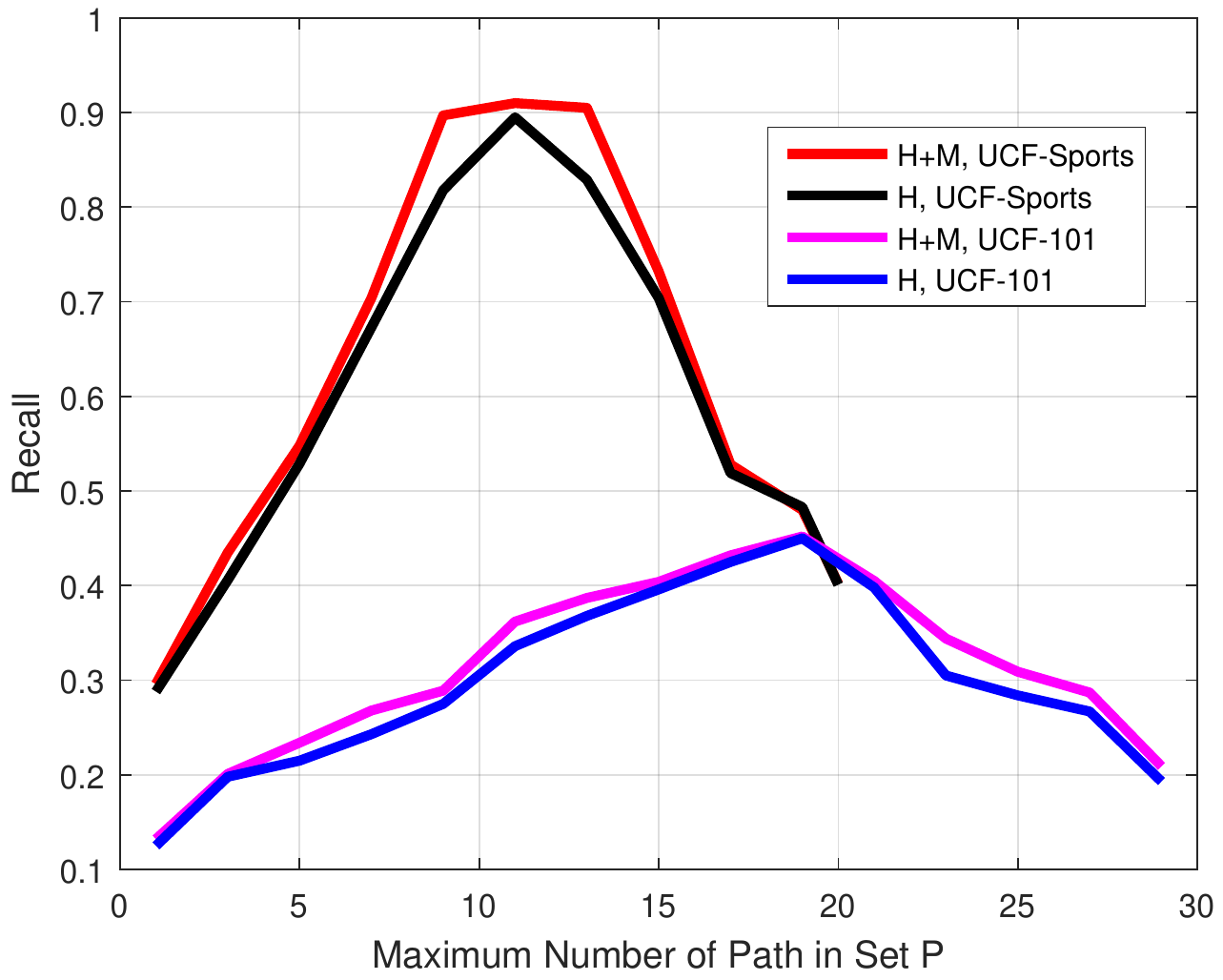}
\caption{Recall vs. maximum number of path in set $\textbf{P}$ under different test settings.}
\label{recall-vs-n}
\end{figure}

\par
For the motion estimation of the bounding boxes, the number of components of GMMs is set to the same as the number of action categories. For constructing GMMs, we use randomly selected 1/3 of the video clips in UCF-101 for training and test on UCF-Sports dataset. While test on UCF-101 dataset, all the video clips of UCF-Sports are used for training. This setting is for a fair comparison with existing non-learning based methods which test on the whole dataset. The number of action paths in a candidate set $\Phi$ is set to $50$ for UCF-Sports and $100$ for UCF-101, as the latter one has a longer duration of action videos on average, and hence may contains more action-path segments. The value of $N$ in Eq.~\ref{MHSCP} (\ie the maximum number of paths in set $\textbf{P}$) is set as $12$ for UCF-Sports and $18$ for UCF-101. For each video clip, we propose at least one path set $\textbf{P}$, while a path set $\textbf{P}$ is generated, the paths $\{p_i, i=1,2,\ldots,N\}$ in $\textbf{P}$ are removed from the candidate set $\Phi$ and the greedy search algorithm is performed again to find a new path set $\textbf{P}^{'}$ until the duration of the longest path $p^{'}_i$ ($p^{'}_i \in \textbf{P}^{'}$) is less than $10$.

\subsection{Analysis of the proposed approach}
We analyze the performance of the proposed approach from different aspects, including the sensitivity of the parameter $N$ (\ie, the maximum number of path in set $\textbf{P}$), the influence of actionness estimation based on human appearance and motion cues, the number of generated proposals and the runtime analysis.
\par Fig.~\ref{recall-vs-n} shows the recall performance of our approach using different actionness estimation schemes by varying the value of $N$. From Fig.~\ref{recall-vs-n}, it can be observed that the proposed approach achieves the best performance when the value of $N$ is in the range $[9, 14]$ on UCF-Sports, and the performance degrades significantly when $N$ is far from this range. The optimal value of $N$ for UCF-101 is larger than that for UCF-Sports. The reason is probably that the video clips of UCF-101 have longer duration than UCF-Sports on average, and thus the action path is more likely to be separated into multiple segments. 

\par We can also observe that the proposed approach using both the human appearance and motion cues for actionness estimation (\ie $\mathrm{H} + \mathrm{M}$) yields better recall performance than that using only the human appearance cue (\ie $\mathrm{H}$) on the two datasets. This demonstrates our initial intuition that employing multiple action-related cues for actionness estimation can help to further improve the performance of action proposal generation.

\par
\begin{figure}[!h]
\centering
\includegraphics[width=0.9\textwidth,height=0.4\textwidth]{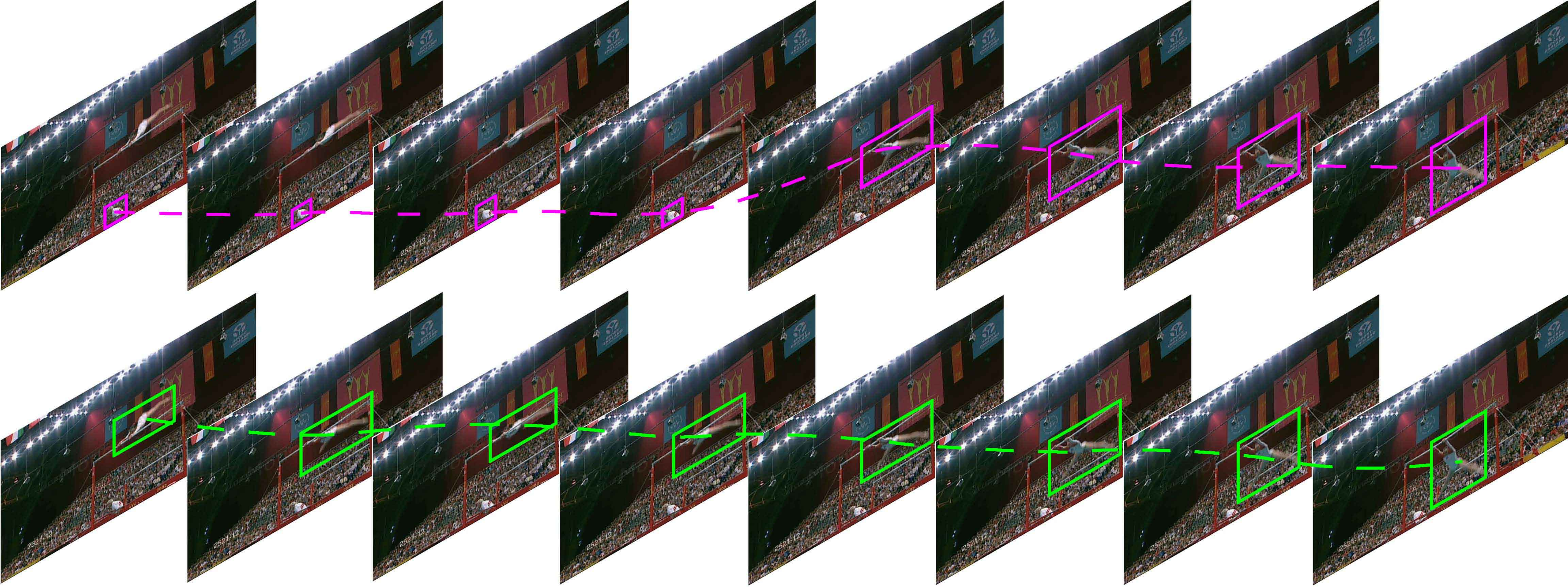}
\caption{Examples of action-path generation results. The 1-st row shows the results obtained from~\cite{yu2015fast} (The action path contains an irrelevant actor within the first few frames); the 2-nd row is our results, where the main actor is correctly tracked.}
\label{action-path}
\end{figure}

As also shown in Fig.~\ref{recall-vs-n}, at the best performance point of recall on UCF-Sports, the number of the generated action proposals of our approach is only $13$, and it is significantly less compared with state-of-the-art methods on the same recall performance level (see Table~\ref{tab1}). The notable improvement is mainly due to the precise human estimation from our fine-tuned Faster-RCNN model, and the modified forward-backward search algorithm for generating candidate set $\Phi$. Compared to~\cite{yu2015fast}, the improved optimization objective leverages appearance similarity among paths for effectively separating different actors. Fig. \ref{action-path} illustrates the improvement by our approach. It can be observed that the action path generated by the proposed approach is correctly associated to the same actor. More examples of the action-proposal generation results on UCF-Sports and UCF-101 are shown in Fig. \ref{resonucfsports} and Fig. \ref{resonucf101}, respectively.

The runtime of the proposed approach includes three parts: (i) spatial actionness estimation: Faster RCNN for human estimation takes around 0.1 seconds per frame (s/f) and GMM-HOF for motion estimation takes around 1 s/f; (ii) temporal action path extraction: the average runtime of this step is 0.09 s/f; and (iii) action path completion takes 0.5 s/f. In summary, the average runtime of the approach is 1.69 s/f. We conduct the runtime analysis on the UCF-Sports dataset with an image resolution of 720 x 404, and based on hardware configurations of an Nvidia Tesla-K80 GPU, 3.4 GHz CPU and 4 GB memory.

\begin{figure*}[!t]
\centering
\includegraphics[height=0.4\textwidth,width=1\textwidth]{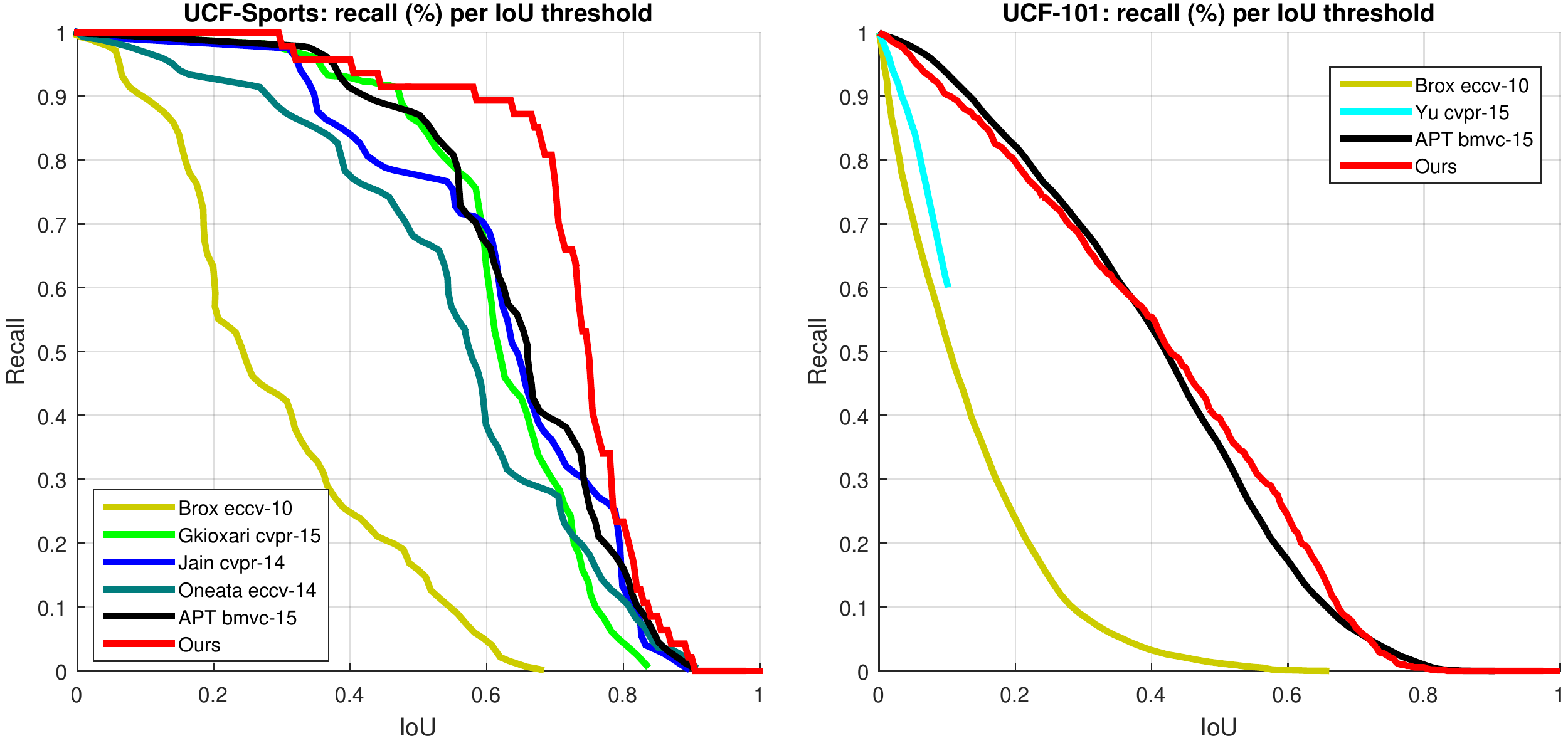}
\caption{Recall vs. IoUs on UCF-Sports and UCF-101 datasets. }
\label{recall-vs-iou}
\end{figure*}

\begin{figure*}[!t]
\centering
\includegraphics[height=0.4\textwidth,width=1.\textwidth]{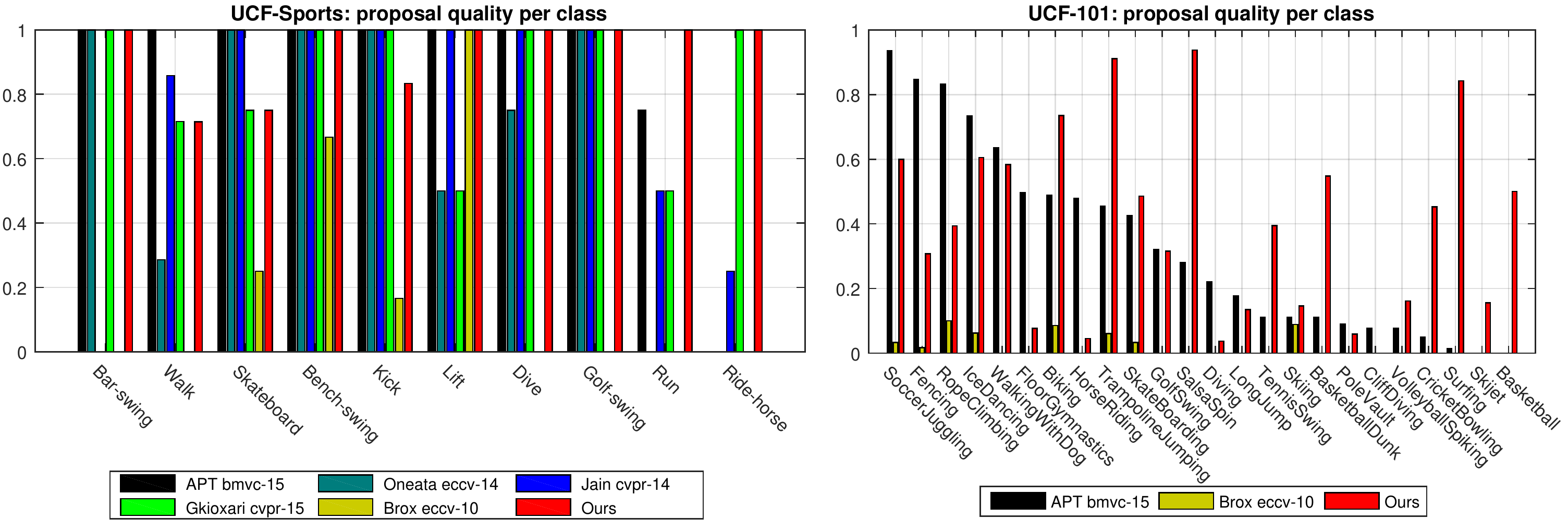}
\caption{Recall performance on each action category on UCF-Sports and UCF-101 datasets.}
\label{qualityperclass}
\end{figure*}

\subsection{Overall Performance}
We compare the performance of the action proposal generation of the proposed approach with state-of-the-art methods on UCF-Sports and UCF-101. We vary the value of IoU ($\eta$) in [0, 1], and plot recall as a function of $\eta$. Fig.~\ref{recall-vs-iou} shows Recall vs. IoU curves of difference approaches. It is clear that our approach obtains a significant performance gap over the state-of-the-art methods on UCF-Sports (The recall of ours is above $0.7$ when $\eta$ is even at $0.7$, while the others are below $0.4$.), and achieves very competitive performance on UCF-101. Fig.~ \ref{qualityperclass} also shows the recall performance for each action category. We can observe that our approach presents superior recall performance on almost all action categories except for few classes (\eg, Walk and Kick) on UCF-Sports, and on UCF-101, ours greatly outperforms the comparison methods on action classes such as Biking, Surfing and TrampolingJumping.

From Fig.\ref{qualityperclass}, it can be also noticed that the performance on UCF-101 is inferior to that on UCF-Sports. The reason is probably because the testing action video clips of UCF-101 have more dynamically and continuously varying size of actors and are more untrimmed than UCF-Sports. Finally, we report the overall performance on the two datasets in Table~\ref{tab1} using several commonly used metrics, including ABO (Average Best Overlap), MABO (Mean ABO over all classes) and the average number of proposals per video. The results further confirm the superior action proposal generation performance achieved by our approach. It should be noted that on the same level recall performance, our approach generates relatively much smaller number of action proposals for each video clip than the other methods, which is especially important for reducing the computational complexity for the follow-up applications such as action recognition and action interaction modeling.

\section{Conclusions}

\begin{table}[t]
\centering
\caption{Quantitative performance comparison of the action proposal generation with state-of-the-art methods with commonly used metrics.}
\footnotesize
\resizebox{1\linewidth}{!}{
\begin{tabular}{lcccc}
\toprule
{} & ABO & MABO & Recall & \#Proposals \\ \hline
UCFSprots \\ \hline
{Brox \& Malik, ECCV 2010 \cite{brox2010object}}  & {29.84} & {30.90} & {17.02} & {4} \\
Jain \etal, CVPR 2014 \cite{jain2014action}       & {63.41} & {62.71} & {78.72} & {1,642} \\
Oneata \etal, ECCV 2014 \cite{oneata2014spatio}     & {56.49} & {55.58} & {68.09} & {3,000} \\
Gkioxari \& Malik, CVPR 2015 \cite{gkioxari2015finding} & {63.07} & {62.09} & {87.23} & {100} \\
APT, BMVC 2015 \cite{van2015apt}                         & {65.73} & {64.21} & {89.36} & {1,449} \\
\textbf{Ours}                           &  \textbf{89.64} & \textbf{74.19} & \textbf{91.49}   &   12     \\ \hline
UCF 101 \\ \hline
Brox \& Malik, ECCV 2010 \cite{brox2010object}    & {13.28} & {12.82} & {1.40} & {3} \\
APT, BMVC 2015 \cite{van2015apt}                         & {40.77} & {39.97} & {35.45} & {2,299} \\
\textbf{Ours}                           &   \textbf{63.76} & \textbf{40.84}  & \textbf{39.64}     &     18   \\ \bottomrule
\end{tabular}
}
\label{tab1}
\end{table}

\begin{figure}[t]
\centering
\includegraphics[width=1\textwidth, height=0.5\textwidth]{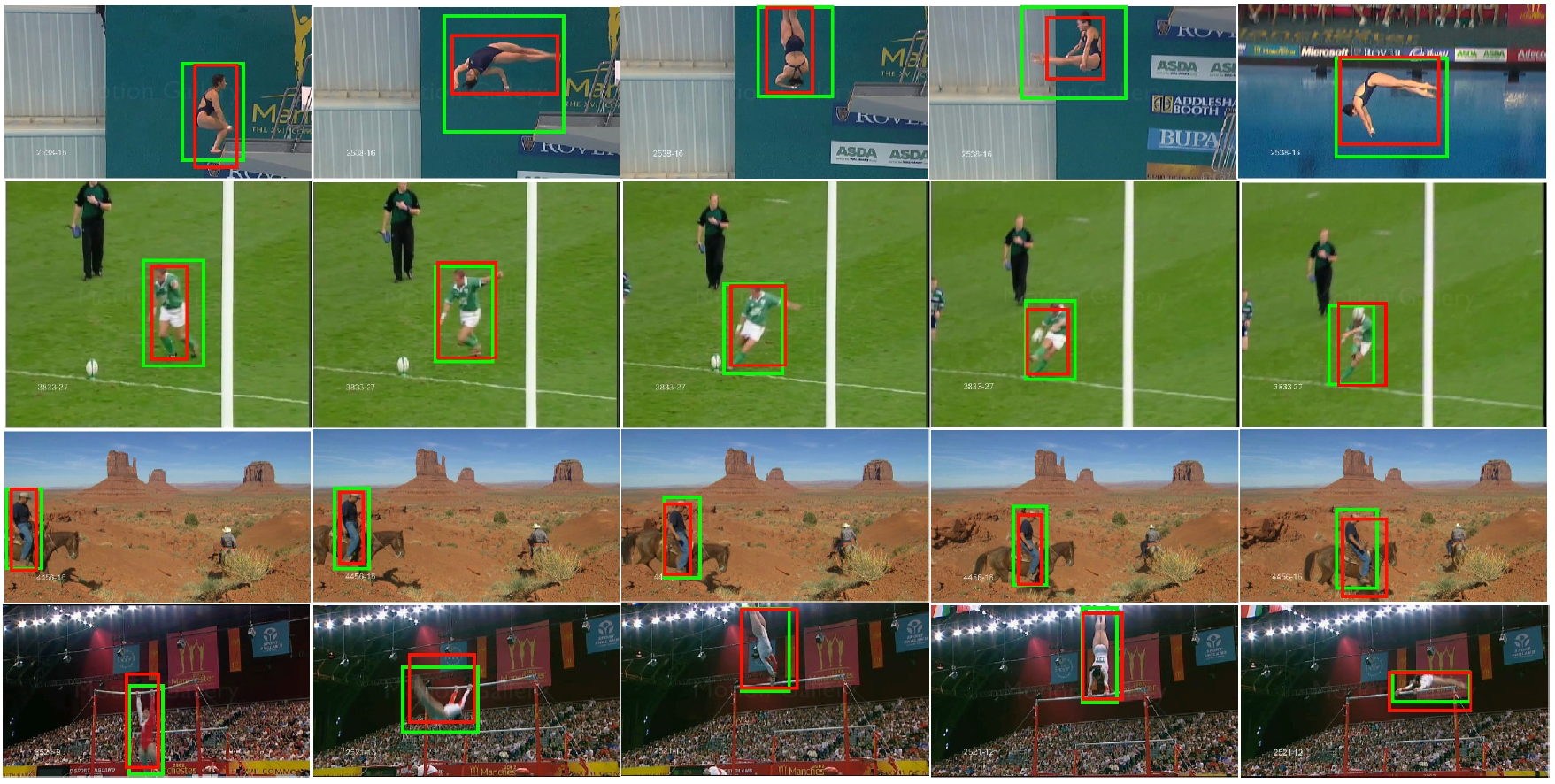}
\caption{Examples of action-proposal generation results on UCF-Sports. The bounding boxes with green and red color are the ground truth and the action proposal, respectively.}
\label{resonucfsports}
\end{figure}

\begin{figure}[!h]
\centering
\includegraphics[width=1\textwidth, height=0.5\textwidth]{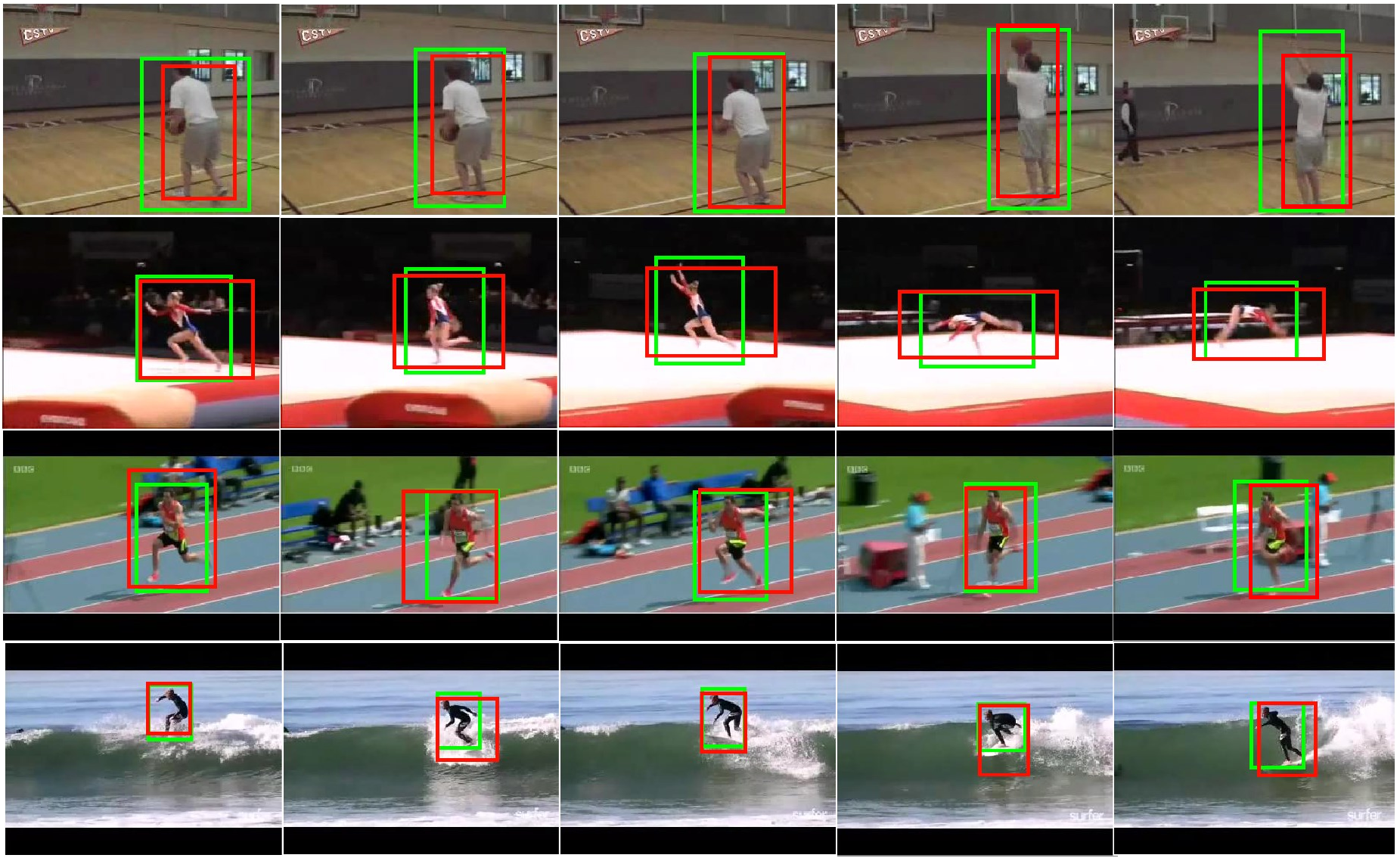}
\caption{Examples of action-proposal generation results on UCF-101. The bounding boxes with green and red color are the ground truth and the action proposal, respectively.}
\label{resonucf101}
\end{figure}

A novel framework for action proposal generation in video has been presented in this paper. Given an unconstrained video clip as input, it generates spatial-temporal continuous action paths. The proposed approach is built upon actionness estimation leveraging both human appearance and motion cues on frame-level bounding boxes, which are produced by a Faster-RCNN network trained on augmented datasets. Then we search spatial-temporal action paths via linking, associating, and tracking the bounding boxes across frames. We formulate the association of action paths belonging to the same actor as a maximum set coverage problem and propose a greedy search algorithm to solve it. Experiments on two challenging datasets demonstrate that our approach produces more accurate action proposals with remarkably less proposals compared with the state-of-the-art approaches. Based on the observation on the experimental results, the proposed approach is especially effective when the action video clip contains only one actor. In the future, we will explore using CNN networks for better actionness estimation of video frames, and consider recurrent neural networks for modeling the action paths for action recognition.

\bibliographystyle{splncs03}
\bibliography{accv2016}
\end{document}